\documentclass[journal]{IEEEtran}

\newcommand{\footnoteline}{\rule{\linewidth}{0.01pt}\vspace*{0.01\baselineskip}}

\usepackage{cite}
\usepackage{amsmath}

\usepackage{hyperref}
\usepackage{url}

\usepackage{graphicx}  %needed to include png, eps figures
\usepackage{multirow}
\usepackage{float}  % used to fix location of images i.e.\begin{figure}[H]

\begin{document}
% paper title
\title{emrQA-msquad: A Medical Dataset Structured with the SQuAD V2.0 Framework, Enriched with emrQA Medical Information}%\small{Creating new dataset in medical field}}

% author names 
\author{{\LARGE Jimenez Eladio and Hao Wu\textsuperscript{a,*}}
\thanks{-----------------------------------------\\
* Corresponding author. Beijing Institute of Technology, No. 5, South Zhongguancun Street, Haidian District, Beijing, 100081, China. E-mail addresses: eladiorjb@gmail.com (J. Eladio), wuhao123@bit.edu.cn (H. Wu).}%
\begin{itemize}
    \item[\textsuperscript{a}] School of Computer Science and Technology, Beijing institute of Technology, Beijing, 100061, China. 
\end{itemize}
\footnoteline%
} 

% make the title area
\maketitle

% As a general rule, do not put math, special symbols or citations in the abstract or keywords.
\begin{abstract}

Machine Reading Comprehension (MRC) holds a pivotal role in shaping Medical Question Answering Systems (QAS) and transforming the landscape of accessing and applying medical information. However, the inherent challenges in the medical field, such as complex terminology and question ambiguity, necessitate innovative solutions. One key solution involves integrating specialized medical datasets and creating dedicated datasets. This strategic approach enhances the accuracy of QAS, contributing to advancements in clinical decision-making and medical research. To address the intricacies of medical terminology, a specialized dataset was integrated, exemplified by a novel Span extraction dataset derived from emrQA but restructured into 163,695 questions and 4,136 manually obtained answers, this new dataset was called emrQA-msquad dataset. Additionally, for ambiguous questions, a dedicated medical dataset for the Span extraction task was introduced, reinforcing the system's robustness. The fine-tuning of models such as BERT, RoBERTa, and Tiny RoBERTa for medical contexts significantly improved response accuracy within the F1-score range of 0.75 to 1.00 from 10.1\% to 37.4\%, 18.7\% to 44.7\% and 16.0\% to 46.8\%, respectively. Finally, emrQA-msquad dataset is publicy available at \href{https://huggingface.co/datasets/Eladio/emrqa-msquad}{https://huggingface.co/datasets/Eladio/emrqa-msquad}.

\end{abstract}


\section{Introduction}

\IEEEPARstart{T}{he} ubiquitous integration of Machine Reading Comprehension (MRC) systems into our daily lives has elevated Question Answering (QA) systems to the forefront of technological progress. As we witness this transition from speculative fiction to an integral part of our daily routines, the inherent challenges within QA systems become increasingly evident. The complexity of these systems necessitates a deep understanding of contextual knowledge to effectively address queries, further complicated by the diverse formulations of questions on the same subject \cite{chang2023survey,wang2023chatcad,mrc_09588221,mrc_10034994,mrc_3617680,mrc_Vietnamese,NEJMra2302038,zhou2023comprehensive,app11125456,thirunavukarasu2023large}.

This research focuses on the critical challenges confronting Medical Question Answering Systems, a domain characterized by inherent variability in clinical information and a scarcity of publicly available medical datasets for training purposes. Challenges include interpreting clinical guidelines, evaluating personalized risks, and navigating the complexities of multidimensional healthcare decision-making \cite{li2022easy,guan2022blockskim,zhang2022greaselm,cao2022coarsegrained,mrc_3560260,chen2018neural,zeng2020,liu2019neural,qiu1906survey}.

To address these issues, our proposal involves the creation of a bespoke dataset in the medical field, seamlessly integrating the structure of SQuAD V2.0 with the rich medical content of the emrQA dataset. This innovative dataset has the potential to enhance the performance of models specifically designed for the medical domain \cite{liu2023deidgpt,nori2023capabilities,sharma2023humanai,mrc_10234662,tapeh2023artificial,lai2023chatgpt}.

\section{Background}
\subsection{EmrQA - Dataset}
EmrQA, a substantial asset in our project, represents a commendable contribution to Question Answering on Electronic Medical Records. This expansive dataset is meticulously curated through an innovative semi-automated generation framework, minimizing expert involvement by repurposing existing annotations from various clinical Natural Language Processing tasks. The generation process involves collecting questions from experts, converting them into templates with placeholders for entities, annotating templates with logical form templates, and utilizing annotations from existing NLP tasks to populate placeholders and generate answers \cite{pampari2018}.

While the medical content within EmrQA holds immense value, it is crucial to acknowledge its non-structured nature. The absence of standardization in the reports hinders the full exploitation of information using traditional Question Answering models. In addition to the intricacies of the medical field's terminology, models tailored for Question Answering struggle to perform optimally on this dataset.

\subsection{SQuAD V2.0 - Dataset}
\begin{figure}[ht]
\centering
\includegraphics[width=0.45\textwidth]{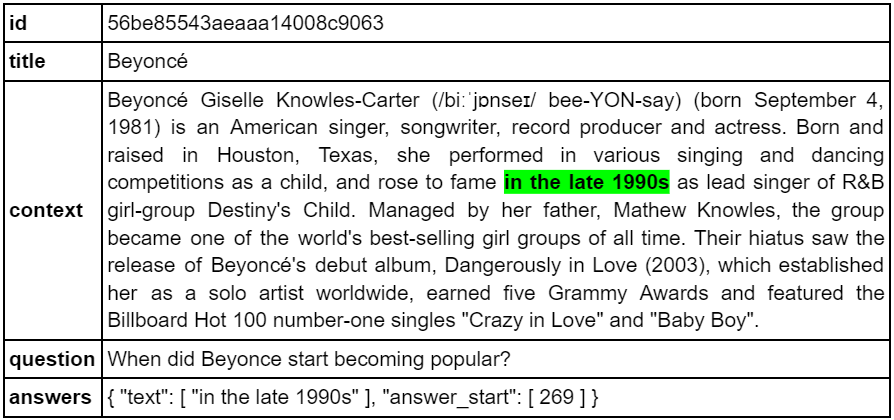}
\caption{Example of SQuAD V2.0 \cite{squad_v2}}\label{fig:example_squadv2}
\end{figure}

SQuAD V2.0 \cite{squad_v2}, Stanford Question Answering Dataset Version 2.0, assumes a central role in our exploration of machine reading comprehension. This dataset, curated by Stanford University, encompasses an extensive repository of over 150,000 questions, standing as a pivotal benchmark in the evaluation and enhancement of question answering systems (Figure \ref{fig:example_squadv2}). Its distinguishing features include a clean, straightforward, and user-friendly structure that has captivated the attention of numerous researchers, establishing itself as a key point of reference across various projects. The meticulously crafted framework of SQuAD V2.0 not only enthralls researchers but also provides a versatile and reliable resource, contributing significantly to the advancement of question answering systems within the domain of machine reading comprehension.

\subsection{Baseline models - BERT, RoBERTa and Tiny RoBERTa}
In our project, before being fine-tuned in the medical field, three distinct models based on Squad V2.0 data serve as the baseline for our Question Answering system within the medical domain. BERT \cite{devlin2019bert}, RoBERTa \cite{liu2019roberta}, and Tiny RoBERTa represent a solid foundation with their proficiency in span extraction and nuanced understanding of general contexts. While these models deliver efficient span extraction and precise information retrieval, their initial performance sets the stage for further enhancement through fine-tuning in the medical domain. This baseline ensures a starting point of robust functionality and accuracy, laying the groundwork for tailored optimization to address the specific challenges and nuances of medical text comprehension.

\begin{figure}[ht]
\centering
\includegraphics[width=0.45\textwidth]{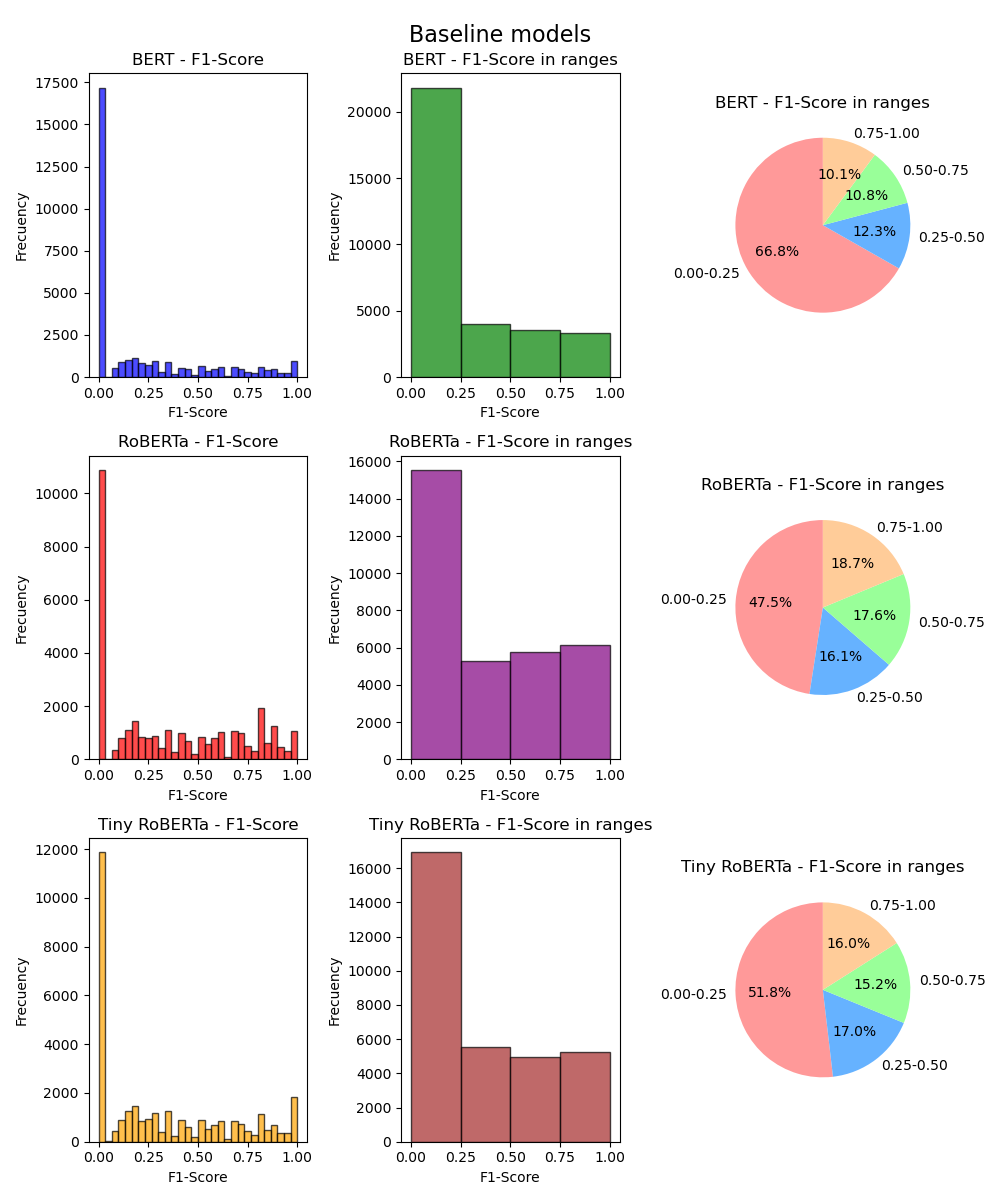}
\caption{Baseline models on emrQA-msquad.}\label{fig:base_lines_models_charts}
\end{figure}

Despite the exceptional performance demonstrated by the BERT, RoBERTa, and Tiny RoBERTa models when operating within general contexts as assessed on SQuAD V2.0, their efficacy significantly waned when applied to a medical dataset. This discrepancy in performance underscores the unique challenges posed by medical text comprehension and highlights the need for specialized fine-tuning to optimize model performance in this domain (Figure \ref{fig:base_lines_models_charts}).

The histograms provide insights into the distribution of responses based on their corresponding F1 score values. We divided the responses into 30 F1 score bins to examine which F1 score values received the highest number of responses (frequency). This analysis allowed us to gain a more detailed understanding of how the baseline models perform poorly within the medical context, as illustrated in Figure~\ref{fig:base_lines_models_charts}.

Additionally, we refined our approach by reducing the number of bins from 30 to 4, each representing a 25\% segment, enabling us to identify the percentage range with the highest F1 score achievements, as shown in Figure~\ref{fig:base_lines_models_charts}. To enhance the visual representation of these findings, we also depicted the latter chart as a pie chart in the same figure, offering a more user-friendly visualization of the data sectors.

An in-depth analysis of the BERT baseline model's performance reveals that a significant portion, specifically 66.8\%, of the responses achieved an F1 score falling within the lower range of 0-0.25. In contrast, only a modest 10.1\% managed to attain a higher F1 score in the range of 0.75-1.00. These results starkly highlight the suboptimal nature of the model when applied to medical context data, pointing towards a notable room for improvement.

Similarly, upon closer examination of the RoBERTa baseline model, it becomes apparent that 47.5\% of the responses obtained an F1 score within the 0-0.25 range, while a slightly larger proportion, 18.7\%, reached an F1 score in the range of 0.75-1.00. Again, this underscores the challenges faced by the baseline model in delivering satisfactory performance within the intricacies of medical context.

Turning our attention to the Tiny RoBERTa baseline model, a noteworthy 51.8\% of responses achieved an F1 score in the lower range of 0-0.25, with only 16.0\% managing a higher F1 score in the 0.75-1.00 range. This consistent pattern underscores the persistent struggles of the baseline model when confronted with the nuances and complexities inherent in medical context data.

In essence, these comprehensive evaluations collectively emphasize the pressing need for refinement and optimization in the baseline models, particularly when addressing the intricate challenges posed by medical domain applications (Figure~\ref{fig:base_lines_models_charts}).

\section{Design and Implementation}

\begin{figure}[ht]
\centering
\includegraphics[width=0.45\textwidth]{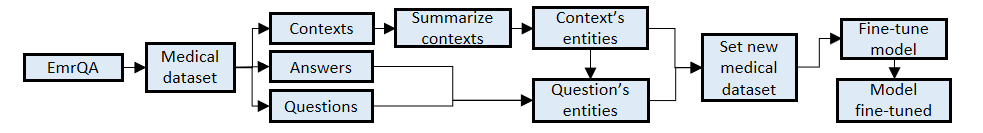}
\caption{EmrQA-msquad scheme.}\label{fig:scheme_span_extraction}
\end{figure}

The dataset creation process was a comprehensive endeavor, initiated by running the emrQA project to procure a medical dataset suitable for our objectives. Leveraging emrQA, we acquired a wealth of contextual information, questions, and medical answers, which proved invaluable. However, to optimize the dataset for our purposes, significant structural modifications were necessary due to the unstructured nature of the original reports. Through a meticulous process, we condensed and structured the reports to concentrate solely on the medical information they contained, while also refining and extracting responses to align with the revised format (Figure \ref{fig:scheme_span_extraction}).

\begin{figure}[ht]
\centering
\includegraphics[width=0.45\textwidth]{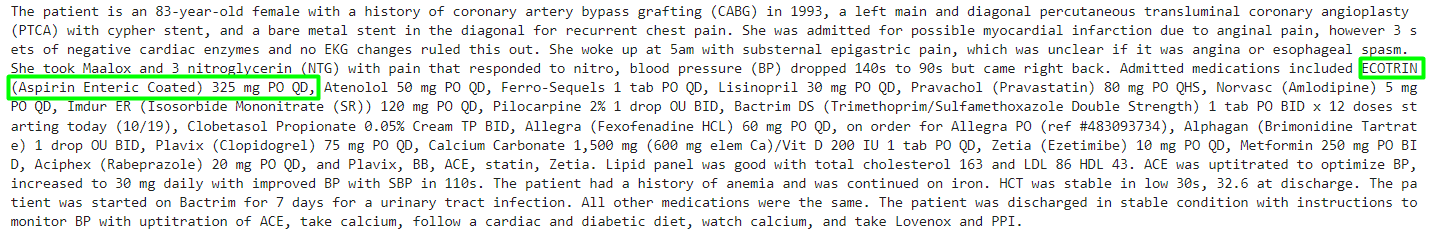}
\caption{Example of emrQA-msquad's medical report with highlighted answer.}\label{fig:emrqa_msquad_report_example}
\end{figure}

\begin{figure}[ht]
\centering
\includegraphics[width=0.45\textwidth]{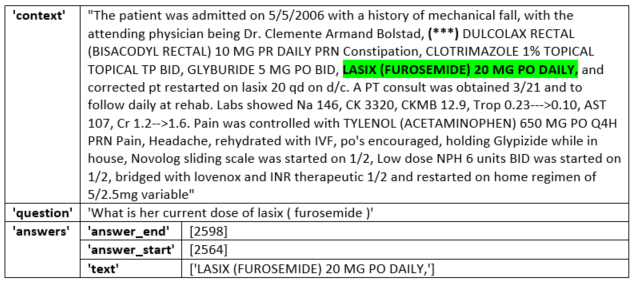}
\caption{Example of emrQA - msquad dataset.}\label{fig:emrqa_msquad_item}
\end{figure}

The contexts in the EmrQA dataset lack structure Figure~\ref{fig:emrqa_msquad_report_example}, making automated summarization unfeasible using code alone. Consequently, summaries were generated using the Large Language Model text-davinci-003 from OpenAI's Davinci model family (Figure \ref{fig:emrqa_msquad_item}). This model was employed to create summaries of medical reports while preserving the content of the answers within the context. However, some answers underwent structural changes, and others were omitted in the summary. This necessitated the creation of a new ground truth for fine tuning the baseline model. Attempts were made to obtain the new ground truth using the text-davinci-003 model, but it proved insufficient for accurate span extraction of answers within the summaries. These models introduced additional information that compromised answer accuracy. As a result, the only viable approach for creating the new ground truth was a time-consuming manual process, where answers were collected one by one.

\begin{figure}[ht]
\centering
\includegraphics[width=0.45\textwidth]{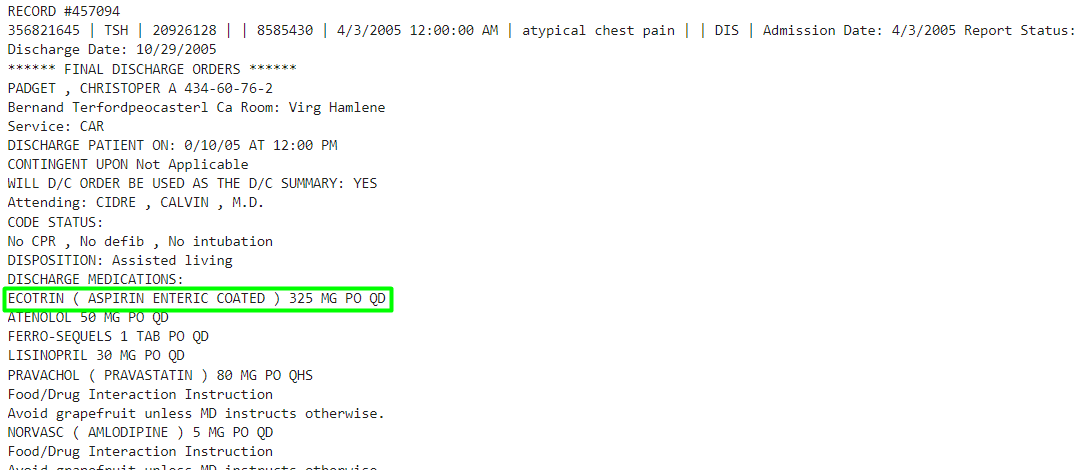}
\caption{Example of emrQA's medical report with highlighted answer.}\label{fig:emrqa_report_example}
\end{figure}

Additionally, the amalgamation of reports with responses in a single prompt led to inputs exceeding 5000 tokens, necessitating careful segmentation into several segments for iterative summarization requests. Optimization measures were applied to minimize repeated responses, reducing the dataset file's dimensionality.

Through meticulous efforts spanning summarization, deduplication of questions and answers, and meticulous manual extraction of contextual details, we've meticulously crafted a dataset that mirrors the format of SQuAD V2.0. Yet, what sets our dataset apart is its exclusive concentration on medical content, a result of a thorough curation process aimed at ensuring its relevance and specificity to the medical domain (Figure \ref{fig:emrqa_msquad_item}).

\section{Results}
Having built our initial Question Answering system using the BERT, RoBERTa, and Tiny RoBERTa models that have excelent results on SQuAD dataset, exact match were 0.71152, 0.79931 and 0.78863 respectively and f1 score were 0.74671, 0.82950, and 0.82035 respectively our goal was to use this model as a baseline and retrain it with medical data to enhance its performance in medical contexts. Knowing that the model had been previously trained with a format provided by SQuAD V2.0, the decision was made to adapt the emrQA-generated database to the SQuAD V2.0 format for the retraining process. This adaptation was crucial to ensure compatibility and leverage the existing structure of the model while incorporating domain specific medical information for improved performance in the medical question-answering domain.

\subsection{emrQA - msquad Dataset}
The \href{https://huggingface.co/datasets/Eladio/emrqa-msquad}{emrQA - msquad dataset} constituted a significant repository of medical information meticulously curated to facilitate the fine-tuning process for Question Answering models. This rich dataset encompassed a diverse array of medical contexts, questions, and corresponding answers meticulously structured and organized, offering an invaluable resource for enhancing the performance and accuracy of machine comprehension within the medical domain (Table \ref{tab:emrqa_msqad_dataset}).

\begin{table}[htbp]
  \centering
  \caption{emrQA-msquad quantity of answers and questions.}\label{tab:emrqa_msqad_dataset}
  \begin{tabular}{|l|c|c|}
  \hline
  \textbf{Dataset} & \textbf{Questions} & \textbf{Answers} \\
  \hline
  emrQA & 210491 & 6636 \\
  \hline
  emrQA-msquad & 163695 & 4136 \\
  \hline  
  \end{tabular}
  \end{table}

The division of this extensive dataset allocates 80\% of its content for training purposes, ensuring that machine learning models can effectively learn from a vast array of medical contexts. The remaining 20\% is strategically reserved for testing, allowing for rigorous evaluation and validation of the models' performance. This meticulously structured dataset is designed to be user-friendly and readily accessible online, facilitating seamless integration into various research and application endeavors within the medical domain. (Table \ref{tab:emrqa_msqad_dataset_object}).

\begin{table}[htbp]
  \centering
  \caption{emrQA-msquad training rows and evaluationg rows.}\label{tab:emrqa_msqad_dataset_object}
  \begin{tabular}{|l|c|c|}
  \hline
  \textbf{Dataset} & \textbf{Train Rows} & \textbf{Validation Rows} \\
  \hline
  new\_medical\_dataset & 130956 & 32739 \\
  \hline
  \end{tabular}
  \end{table}

\subsection{Fine tuned models - BERT, RoBERTa and Tiny RoBERTa}

During the fine-tuning process, the custom database, derived from the emrQA paper and formatted to align with the SQuAD V2.0 structure, played a crucial role. This meticulous adaptation ensured compatibility with the chosen BERT architecture, specifically the BERT, RoBERTa, and Tiny RoBERTa models. The utilization of Google Colab and Jupyter tools provided a collaborative and interactive environment for seamless development, experimentation, and model refinement.

The fine tuned models (Table \ref{tab:hyperparameters}) incorporates knowledge from both SQuAD and emrQA datasets, leveraging their respective strengths in question-answering contexts and medical domain understanding. This amalgamation contributes to a model that is not only adept at handling general question-answering tasks but is also tailored to excel in the intricacies of medical queries.

The fine-tuned model stands as a testament to the synergistic integration of domain-specific data, advanced language models, and collaborative development tools, representing a robust solution for medical question-answering applications.

\begin{table}[htbp]
  \centering
  \caption{Training hyperparameters of fine tuned models on emrQA-msquad}\label{tab:hyperparameters}  
  \begin{tabular}{|c|c|}%{\textwidth}{|X|X|}
    \hline
    \textbf{Hyperparameters} & \textbf{Value} \\
    \hline
    max\_length & 512 \\
    \hline
    learning\_rate & 2e-5 \\
    \hline
    num\_train\_epochs & 3 \\
    \hline
    weight\_decay & 0.01 \\
    \hline
    doc\_stride & 172 \\
    \hline
    batch\_size & 16 \\
    \hline
  \end{tabular}
\end{table}

Analyzing the outcomes of Span Extraction models when applied to medical contexts, considering metrics such as Exact Match, Precision, Recall, BLEU, ROUGE and F1-score (Tables \ref{tab:span_metrics1} and \ref{tab:rouge_table}). The results obtained from the Baseline model emphasize the critical role of the model's training environment. This is evident in the significantly poor and almost negligible results observed when this model is employed with medical context data. In contrast, the fine-tuned model, which was customized using the medical database we meticulously generated from scratch, tailored explicitly to the content of emrQA, produced markedly superior results when compared to the reference medical dataset.

These results underscore the paramount importance of fine-tuning models for specific tasks to achieve superior performance. Continuous refinement and exploration of techniques hold the potential for achieving enhanced metrics across all Machine Reading Comprehension tasks.

\begin{table}[ht]
  \centering
  \caption{Metrics results of Fine tuned models}\label{tab:span_metrics1}
  \begin{tabular}{|l|c|c|c|c|c|c|}
    \hline
    \textbf{MRC Tasks} & \textbf{Exact Match} & \textbf{Precision} & \textbf{Recall} & \textbf{F1-Score} & \textbf{BLEU} \\
    \hline
    %Cloze test (1 epoch) & 0.4162 & 0.48 & 0.42 & 0.43 \\
    BERT BLM & 0.0283 & 0.01 & 0.00 & 0.00 & 0.071 \\
    BERT FTM & 0.3128 & 0.31 & 0.31 & 0.30 & 0.139 \\
    RoBERTa BLM & 0.0296 & 0.02 & 0.01 & 0.01 & 0.112 \\
    RoBERTa FTM & 0.4148 & 0.41 & 0.41 & 0.41 & 0.182 \\  
    Tiny RoBERTa BLM & 0.0529 & 0.02 & 0.01 & 0.01 & 0.116 \\
    Tiny RoBERTa FTM & 0.3962 & 0.39 & 0.39 & 0.38 & 0.251 \\      
    \hline
  \end{tabular}
\end{table}

In Table \ref{tab:span_metrics1}, to streamline readability and facilitate data analysis, we have opted to utilize the abbreviations "BLM" to represent the "Baseline Model" and "FTM" to denote the "Fine-tuned Model." These abbreviations offer a succinct yet informative representation of the respective model categories, aiding in the interpretation of the presented metrics.

\begin{table}[ht]
  \centering
  \caption{Metrics rouge results of Fine tuned models}
  \label{tab:rouge_table}
  \begin{tabular}{|c|c|c|c|c|c|c|}
    \hline
    \multirow{2}{*}{\textbf{ROUGE}} & \multicolumn{3}{c|}{\textbf{BERT Baseline model}} & \multicolumn{3}{c|}{\textbf{BERT Fine tuned model}} \\
    \cline{2-7}
    & \textbf{F1-score} & \textbf{Precisión} & \textbf{Recall} & \textbf{F1-score} & \textbf{Precision} & \textbf{Recall} \\
    \hline
    ROUGE-1 & 0.207 & 0.367 & 0.166 & 0.466 & 0.704 & 0.423 \\
    ROUGE-2 & 0.146 & 0.253 & 0.118 & 0.389 & 0.447 & 0.370 \\
    ROUGE-L & 0.211 & 0.365 & 0.167 & 0.466 & 0.703 & 0.423 \\
    \hline
    \multirow{2}{*}{\textbf{ROUGE}} & \multicolumn{3}{c|}{\textbf{RoBERTa Baseline model}} & \multicolumn{3}{c|}{\textbf{RoBERTa Fine tuned model}} \\
    \cline{2-7}
    & \textbf{F1-score} & \textbf{Precisión} & \textbf{Recall} & \textbf{F1-score} & \textbf{Precision} & \textbf{Recall} \\
    \hline
    ROUGE-1 & 0.331 & 0.578 & 0.265 & 0.512 & 0.748 & 0.480 \\
    ROUGE-2 & 0.253 & 0.439 & 0.202 & 0.446 & 0.472 & 0.437 \\
    ROUGE-L & 0.338 & 0.574 & 0.271 & 0.516 & 0.747 & 0.484 \\
    \hline 
    \multirow{2}{*}{\textbf{ROUGE}} & \multicolumn{3}{c|}{\textbf{Tiny RoBERTa Baseline model}} & \multicolumn{3}{c|}{\textbf{Tiny RoBERTa Fine tuned model}} \\
    \cline{2-7}
    & \textbf{F1-score} & \textbf{Precisión} & \textbf{Recall} & \textbf{F1-score} & \textbf{Precision} & \textbf{Recall} \\
    \hline
    ROUGE-1 & 0.304 & 0.536 & 0.248 & 0.527 & 0.676 & 0.495 \\
    ROUGE-2 & 0.223 & 0.375 & 0.184 & 0.481 & 0.541 & 0.460 \\
    ROUGE-L & 0.309 & 0.534 & 0.253 & 0.533 & 0.675 & 0.501 \\
    \hline   
  \end{tabular}
\end{table}

\begin{figure}[ht]
\centering
\includegraphics[width=0.45\textwidth]{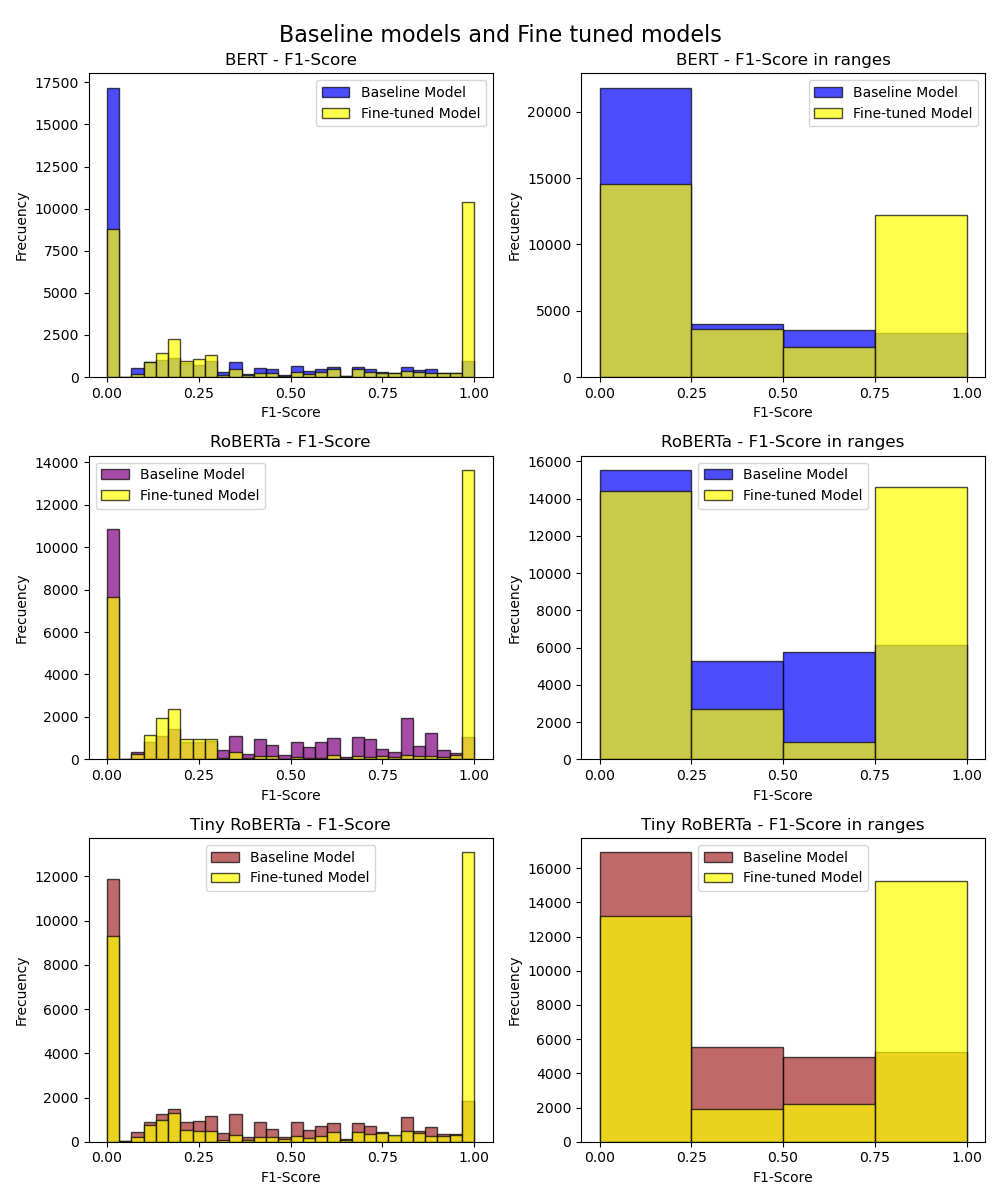}
\caption{BERT, RoBERTa and Tiny RoBERTa on emrQA-msquad.}\label{fig:baseline_fine_tuned_models_charts}
\end{figure}

The histograms provide insights into the distribution of responses based on their corresponding F1 score values. We divided the responses into 30 F1 score bins to examine which F1 score values received the highest number of responses (frequency). This analysis allowed us to gain a more detailed understanding of how the fine-tuned model performs better than baseline within the medical context, as illustrated in Figure~\ref{fig:baseline_fine_tuned_models_charts}.

Additionally, we refined our approach by reducing the number of bins from 30 to 4, each representing a 25\% segment, enabling us to identify the percentage range with the highest F1 score achievements, as shown in Figure~\ref{fig:baseline_fine_tuned_models_charts}. To enhance the visual representation of these findings, we also depicted the latter chart as a pie chart as the last figure, offering a more user-friendly visualization of the data sectors.

The performance analysis reveals intriguing insights into the behavior of the baseline models, particularly BERT, RoBERTa, and Tiny RoBERTa. Each model showcases distinctive trends in response distribution across different F1 score ranges, shedding light on the impact of model retraining within a specialized context (Figure \ref{fig:baseline_fine_tuned_models_charts}).

Firstly, focusing on the BERT baseline model, there is a significant shift in response distribution. The proportion of responses falling within the lower F1 score range of 0-0.25 witnesses a substantial decrease, plummeting from 66.8\% to 44.7\%. Conversely, there is a remarkable surge in responses within the higher F1 score range of 0.75-1.00, skyrocketing from 10.1\% to 37.4\%. This marked transformation underscores the efficacy of retraining the model within a specific domain, where fine-tuning processes are tailored to enhance performance. The meticulous adjustments made during fine-tuning, particularly within the medical domain, contribute significantly to this observed improvement.

Similarly, the RoBERTa baseline model exhibits a parallel trend, albeit with its nuances. The percentage of responses within the 0-0.25 F1 score range decreases from 47.5\% to 44.1\%, indicating a notable decline. Conversely, there is a notable uptick in responses within the 0.75-1.00 F1 score range, rising from 18.7\% to 44.7\%. This pattern reinforces the significance of retraining models within specific contexts, a strategy proven effective in optimizing performance. The tailored fine-tuning process ensures that the model adapts seamlessly to the intricacies of the medical domain, thus yielding substantial improvements in response accuracy and relevance.

Likewise, the Tiny RoBERTa baseline model showcases a comparable pattern of response distribution. The decrease in responses within the 0-0.25 F1 score range, from 51.8\% to 40.6\%, highlights the model's adaptability to the retraining process. Concurrently, there is a notable increase in responses within the 0.75-1.00 F1 score range, climbing from 16.0\% to 46.8\%. This observation underscores the transformative impact of fine-tuning within a specialized domain, where the model's proficiency in extracting relevant information from medical contexts is significantly enhanced. The iterative refinement process ensures that the model evolves to deliver accurate and contextually relevant responses, thereby augmenting its utility within medical applications.

It can gain better insight into how the overlapping results in the histograms illustrate the substantial enhancement achieved by the fine-tuned models compared to the baseline models. This improvement is characterized by more precise and accurate outcomes, as evidenced by the reduction in responses with almost negligible results and a significant increase in responses within the top 25\% F1 score accuracy range (Figure \ref{fig:baseline_fine_tuned_models_charts}).

\section{Discussion and Summary}
The most notable accomplishment is the development of a novel, refined dataset tailored for the medical domain. Drawing insights from the EmrQA dataset and adopting the structure of the SQuAD V2.0 dataset, this newly created dataset encompasses 253 contexts, 163,695 questions, and 4,136 answers, offering a clean and optimized resource for further exploration.

The culmination of this research effort has yielded a highly effective Question Answering (QA) system with integrated entity classification, tailored explicitly for medical data. A key revelation lies in the substantial enhancements achieved by the span extraction prediction models when operating in medical contexts. There are improvements in BERT, Roberta and Tiny Roberta models, Exact Match from 0.0283 to 0.3128, 0.0296 to 0.4148 and 0.0529 to 0.3962, respectively. The F1 score, a crucial metric assessing model performance, witnessed a significant rise from 10.1\% to 37.4\%, 18.7\% to 44.7\% and 16.0\% to 46.8\%, also respectively, on the critical range of 0.75 to 1.0. 

This notable progress signifies the model's heightened capability to accurately extract pertinent information from medical texts, showcasing its proficiency in comprehending nuanced and domain-specific content. Furthermore, a discernible shift in prediction distribution is evident, as the prevalence of predictions in the 0.00-0.25 range diminished in each fine tuned model. Those shifts not only indicates improved precision in low-confidence predictions but also underscores the model's adaptability to the intricacies of medical language. 

%\bibliographystyle{ieeetr}
%\bibliography{bibliography}
%\printbibliography
\bibliographystyle{ieeetr}
\bibliography{bibliography}

\end{document}